\documentclass[letterpaper, 10 pt, conference]{IEEE/ieeeconf}  
\usepackage[utf8]{inputenc}
\usepackage{amsmath, amssymb, graphicx, xfrac, amsfonts, amsbsy, algorithm, color, subfig, cite, url, multirow}
\usepackage[noend]{algpseudocode}
\usepackage[section]{placeins}
\usepackage{tabularx}
\let\labelindent\relax
\usepackage{supertabular,enumitem}
\DeclareMathOperator*{\argmax}{argmax}
\usepackage[countmax]{subfloat}

\IEEEoverridecommandlockouts                              

\title{\LARGE \bf
Aerial Imagery based LIDAR Localization for Autonomous Vehicles 
}
\date{February 2020}

\author{Ankit Vora$^{1}$, Siddharth Agarwal$^{1}$, Gaurav Pandey$^{2}$ and James McBride$^{2}$
\thanks{Ankit Vora and Siddharth Agarwal are co-first authors}
\thanks{$^{1}$Ankit Vora and Siddharth Agarwal are with Ford AV LLC, Dearborn, MI, USA
        {\tt\small \{avora3,sagarw20\}@ford.com}}%
\thanks{$^{2}$Gaurav Pandey and James McBride are with Ford Motor Company, Dearborn, MI, USA
        {\tt\small \{gpandey2,jmcbride\}@ford.com}}%
}

\begin{document}

\maketitle
\thispagestyle{empty}
\pagestyle{empty}


\begin{abstract}
	
This paper presents a localization technique using aerial imagery maps and LIDAR based ground reflectivity for autonomous vehicles in urban environments. Traditional localization techniques using LIDAR reflectivity rely on high definition reflectivity maps generated from a mapping vehicle. The cost and effort required to maintain such prior maps are generally very high because it requires a fleet of expensive mapping vehicles. In this work we propose a localization technique where the vehicle localizes using aerial/satellite imagery, eradicating the need to develop and maintain complex high-definition maps. The proposed technique has been tested on a real world dataset collected from a test track in Ann Arbor, Michigan. This research concludes that aerial imagery based maps provides real-time localization performance similar to state-of-the-art LIDAR based maps for autonomous vehicles in urban environments at reduced costs.

\end{abstract}


\section{INTRODUCTION}

Localization is at the core of any autonomous vehicle system. Various sub-systems like perception, planning and control rely on precise vehicle location. The localization requirements for autonomous systems have evolved over time. A vehicle with driver assist features requires lower lane level accuracy as compared to level 4 \& 5 autonomous vehicles, which need sub decimeter accuracy\cite{Reid_2019}. Along with the requirements, the sensors that are used for localization have evolved as well. GPS was one of the first sensors to be used for global localization. However, use of GPS alone cannot guarantee consistently low error in all environments unless high-grade IMUs are used\cite{ndjeng2011low}. The IMUs in general suffer from drift and biases which can be corrected by Differential GPS (DGPS) or Real Time Kinematics (RTK)\cite{swiftnav} corrections. In open sky conditions, these corrections provide centimeter level accuracy but suffer from multi path errors in downtown and urban canyon environments\cite{obst2012multipath}. As a result, autonomous vehicles rely on map based localization techniques which involves matching real time detected features with a prior map\cite{Levinson.Thrun2007}. 

The prior maps contain various types of data like ground reflectivity\cite{Levinson.Thrun2010}, 3D point clouds, road markings\cite{ranganathan}, etc. In order to create highly accurate high definition maps, companies have to maintain survey vehicles with high grade sensors like LIDARs and cameras. These vehicles collect large amount of data which is post processed with human in the loop. As a result, these maps are expensive and needs to be updated on frequent basis.

\begin{figure}[t]
	\centering
    \subfloat[Orthographic view of our online localization]{\includegraphics[width=0.90\linewidth, trim=0 90 0 0, clip]{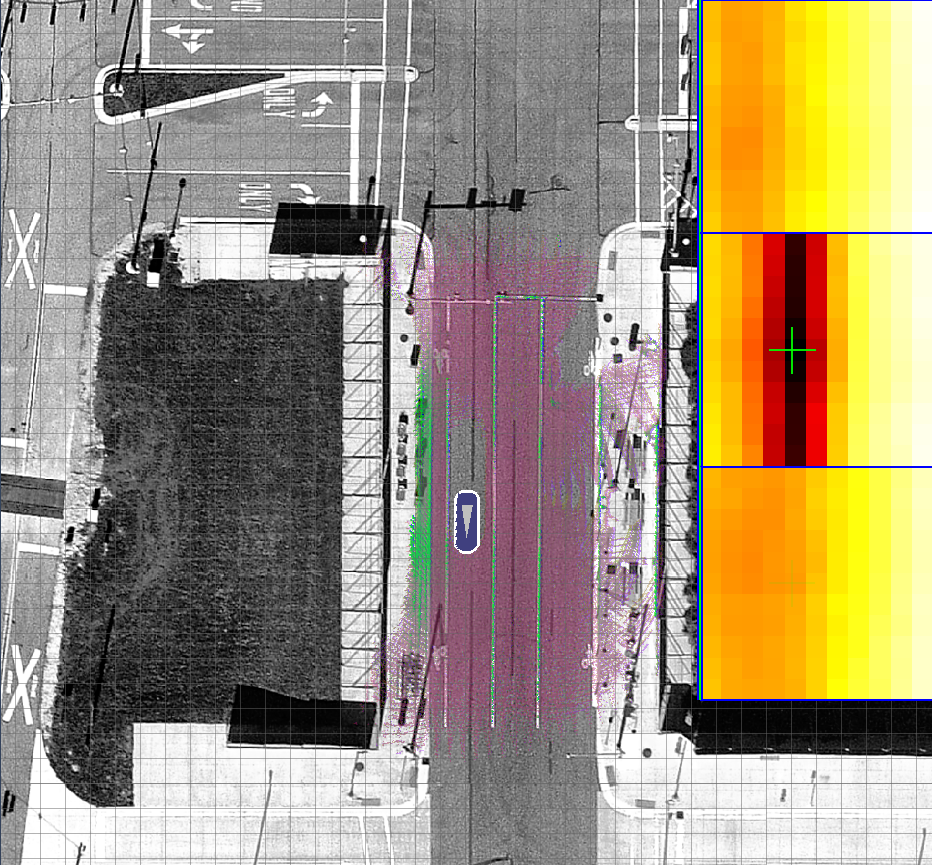}}
	\vspace{0mm}
    \subfloat[Perspective view of the vehicle at the same instant]{\includegraphics[width=0.90\linewidth, trim=0 30 0 100, clip]{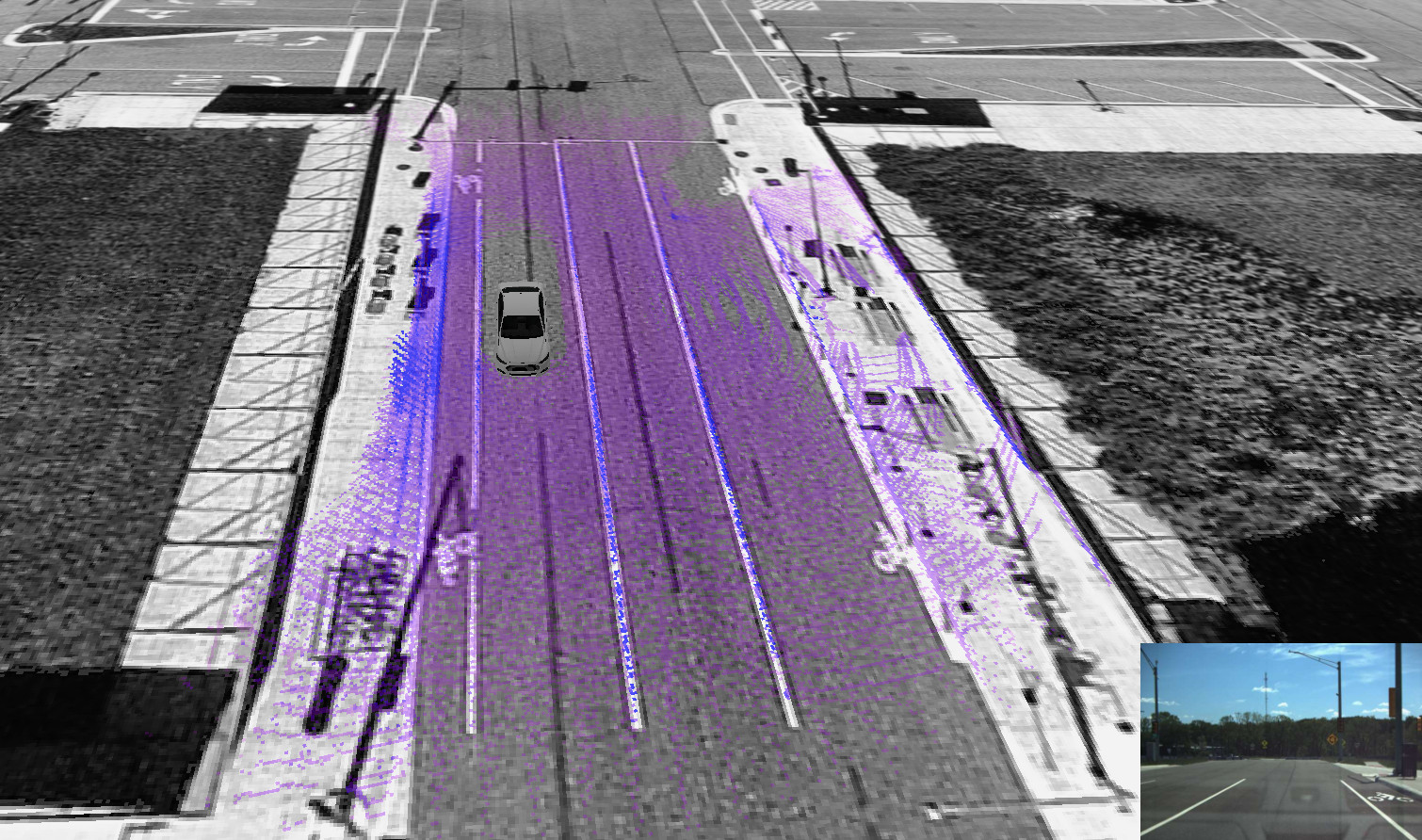}}

	\caption{ Visualization of proposed localization system. The vehicle localized by matching live LIDAR reflectivity against the aerial imagery based prior map using NMI.} 
	\label{fig:filterGUI}
\end{figure}

In this paper, we propose a system which exploits publicly available aerial imagery to create a prior map. This is a relatively cost effective solution as there is no need to maintain a fleet of survey vehicles. We then localize a vehicle by comparing live LIDAR ground reflectivity overlayed on several candidates, seeking to maximize normalized mutual information (NMI) and filtering it using an EKF framework(as outlined in Fig. \ref{fig:filterGUI}).
The key contributions of our paper are:

\begin{enumerate}[noitemsep,partopsep=0pt,topsep=0pt,parsep=0pt]
\item[(i)]  We present a multi-modal localization approach which allows us to use aerial imagery as a prior map to match against live LIDAR ground reflectivity.
\item[(ii)]  We compare our results with state-of-the-art ground reflectivity based localization technique and demonstrate comparable performance.
\item[(iii)] We present experimental results to show that this technique can be used with low resolution satellite maps, thereby enabling a novel and cost effective localization technique for Level 2 \& above autonomous vehicles. 
\end{enumerate}

\section{RELATED WORK}

Localization is a challenge the research community is trying to solve for years now. In 2005, US Department of Defense organized a competition to advance the algorithms and techniques in the field of autonomous vehicles. Most of the teams relied on using GPS, IMU and wheel odometry in order to localize the vehicles globally\cite{DARPA}. The challenge in using the above mentioned sensors is that these have large errors (upto 5m) or drift over time which causes degraded performance of the overall localization system.

In order to solve the above challenge, Levinson et al.\cite{Levinson.Thrun2007} proposed a novel way of localizing an autonomous vehicle using GPS, IMU, wheel odometry and LIDAR based map matching. The live LIDAR measurements are compared to a SLAM\cite{slamBosse, slamThrun} corrected prior map  and tracked using particle filter. By doing so, the authors are able to get sub-decimeter level of accuracy which is a magnitude lesser than any of the previously established localization techniques. 

Ground plane reflectivity is not the most robust features to detect as it is prone to occlusion. As a result, Wolcott et al.\cite{Wolcott17} proposed a way to model the world as a multiresolution map of mixture of Gaussians. Along with the ground plane reflectivity, these gaussians also take into account the height distribution of the environment. Real time matching between LIDAR measurements and the gaussian mixtures using NMI gives the authors accuracies of 0.15m. In another line of work, Rohde et al.\cite{Rohde} used a multi layered approach to slice 3D point cloud map into 2D projections and compare against LIDAR data using AMCL\cite{FoxAMCL}. All the map matching techniques covered so far rely on an accurate prior map that needs to be built using survey grade sensors like GPS, IMU, LIDARs. An alternate to creating these maps is using publicly available aerial imagery.

The use of aerial or satellite imagery is not new to robotics and autonomous vehicles community. Dogruer et al.\cite{dogruer2010outdoor} used satellite images to localize an autonomous robot using Monte Carlo localization\cite{FoxAMCL}. Similarly, Senlet et al.\cite{SenletRobotAerial} used stereo images to create an orthographic view of the environment in front of the robot. These images when matched with aerial imagery based maps furnishes accurate robot localization even without wheel odometry. These methods were improvised upon and used by Noda et al.\cite{NodaInVehicle} to localize an autonomous vehicle instead of a robot. Further advancing the work in the field of aerial imagery based localization of autonomous vehicles, Veronese et al.\cite{Veronese} used real time LIDAR measurements to match against aerial imagery based prior map. These images are low resolution which contributed to feature blurring and as a result, the authors obtained error of 0.89m over a 6.5km trajectory. Since the authors did not have a method to compute ground truth trajectory, they calculated the error based on the positions of known landmarks in the global frame. 

We present a system to localize an autonomous vehicle using on-board sensors like GPS, IMU and LIDAR. We use an Extended Kalman Filtering framework to track the vehicle's pose over time. A simple constant velocity based vehicle motion model is used to predict vehicle pose which is then corrected by the NMI based map matching between aerial imagery based prior map and live LIDAR based ground reflectivity map. In our work, we compare the error in the localization system with SLAM\cite{slamBosse} corrected trajectory and state-of-the-art ground reflectivity based localization results\cite{Levinson.Thrun2007}. We also provide evidence that the magnitude of localization error does not change considerably as we go from high resolution images to low resolution images to create the prior map.

\section{PROPOSED APPROACH}

Our localization framework is based on using measurements from 3D LIDAR to create a local ground reflectivity grid map. As the vehicle moves, localization is run online by matching the current local map with the prior map\cite{Levinson.Thrun2007,Levinson10}. We refer to our aerial map based method as Aerial Localization - \(AL\) and the state-of-the-art method as LIDAR Ground Reflectivity - \(LGR\)  for benchmarking purposes.

\subsection{Global Prior Map}
\label{ssec:globalMap}

Our global map is based on high resolution aerial imagery provided by Maxar Technologies\cite{maxar}. Maxar partnered with Nearmap\cite{nearmap} to provide 8 cm aerial imagery using an aircraft equipped with an under-mounted camera system\cite{nixon2013systems}, a specialized airborne GPS system and an inertial measurement unit. Aerial imagery is acquired using a specialized flight path and typically done during optimal weather and sun conditions.  Once the imagery is acquired, it goes through a series of post-processing to tie all the images together and correct for the Earth’s terrain. These processes are called bundle adjustment\cite{kume2015bundle} and orthorectification\cite{ortho}. These processes correct the terrain distortion in images that results from variations in the surface of the Earth and tilt of the sensor being used to collect the data. This process allows accurate information to be gathered from images such as distances, angles, and positions. The huge advantage of this imagery is to be able to see everything from a top-down vantage point; we can see every road intersection, driveway, parking lot, etc. (i.e. everywhere a vehicle needs to navigate).

\begin{figure*}[t]
    \centering
	\subfloat[AL Prior Map - RGB]{\includegraphics[width=0.312\linewidth, trim=0 0 0 80, clip]{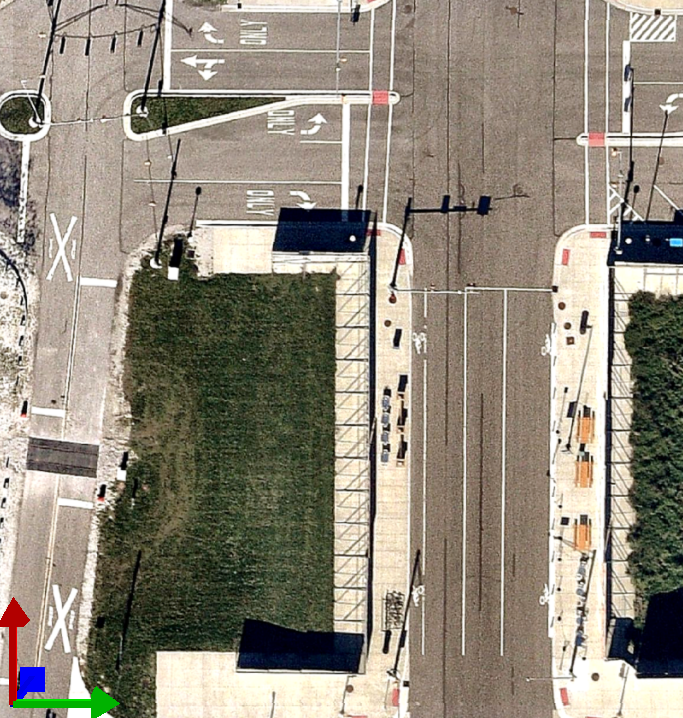}}
	\hspace{3mm}
	\subfloat[AL Prior Map - Grayscale]{\includegraphics[width=0.312\linewidth, trim=0 0 0 77, clip]{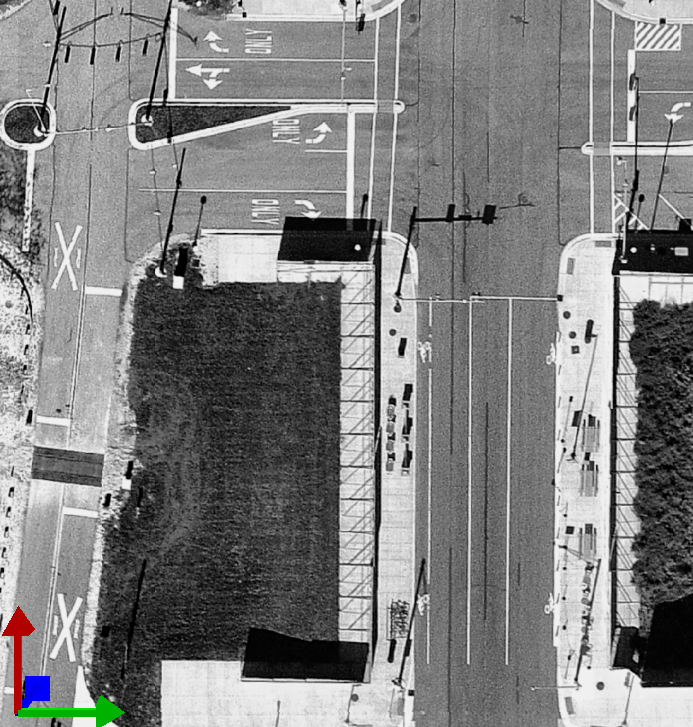}}
	\hspace{3mm}
	\subfloat[Ground Reflectivity Prior Map]{\includegraphics[width=0.312\linewidth, trim=0 0 0 80, clip]{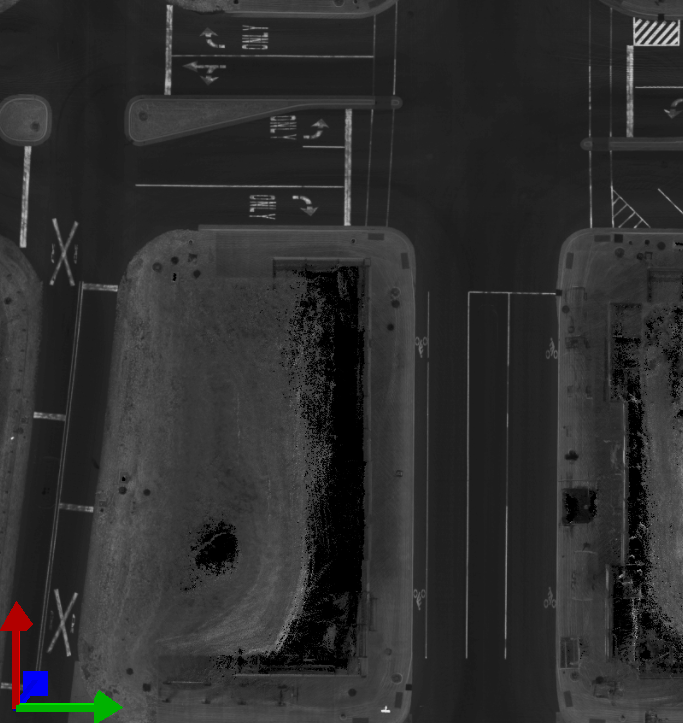}}
	\caption{(a) Aerial imagery based prior map with RGB information. (b) Aerial imagery based prior map converted to grayscale, used for localization. (c) LIDAR ground reflectivity based prior map for benchmarking against state-of-the-art localization technique\cite{Levinson.Thrun2010}. Each image represents one 64m x 64m tile within a larger prior map.} 
	\label{fig:priorMaps}
\end{figure*}

These aerial images are tiled such that each tile represents a 64m x 64m area in the real world. The nominal operating cell resolution is 8cm which implies an image size of 800 x 800 pixels for each tile. These images are then converted to grayscale grid map with each cell in the grid map having an \(X,Y\) coordinate with respect to the map origin. This process helps create a 2D grid representation out of aerial images as shown in fig.~\ref{fig:priorMaps}. The origin of this grid is chosen arbitrarily and stored along with it's latitude and longitude. This information is used to convert a GPS coordinate (Lat, Lon) to an X,Y coordinate in the linearized global frame as required in section \ref{ssec:filter}.

We also create 16 cm and 32 cm maps to mimic the highest resolution satellite images that are legally allowed. For satellite data, licenses to operate satellites are regulated by the Federal Government. Satellite image is a global complement and viable alternative to aerial imagery, particularly for countries with formidable legal, geographic, or geopolitical barriers to aerial-based imagery collection, or where fast delivery and regular, cost-effective updates are major requirements. Over 100 countries around the world fall into that category\cite{maxar}. 

\subsection{Local Map}
\label{ssec:localMap}

\begin{figure}[h]
	\centering
	\captionsetup{justification=centering}
	\subfloat{\includegraphics[width=0.81\linewidth]{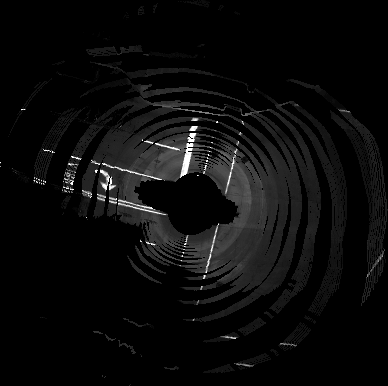}}\\
	\caption{Local Ground Reflectivity Map}
	\label{fig:localMap}
\end{figure}

Localization is run online by matching a local grid map around the vehicle with the prior map along the x-y plane. The cell values within this local grid which we chose to be of size $40 \times 40$ m represent the ground reflectivity of the environment. The values range between [0,255] and represent the mean of all the laser returns obtained from the corresponding cell. The cell resolution of the grid map is matched to that of the prior map. The local grid map is computed and updated online from the accumulated LIDAR points which are motion compensated using inertial measurements and ground segmented. An example of local ground reflectivity grid map is shown in Figure \ref{fig:localMap}.

\subsection{Normalized Mutual Information for Registration}
\label{ssec:NMIregistration}

In the field of medical imaging, NMI has been widely used to match images from different sources\cite{MaesImageRegistration}. Theoretically, NMI provides a measure of similarity between two independent distributions. In our work, we compare the similarity between aerial imagery based prior map and local LIDAR reflectivity. We assume a planar GPS constraint of zero height which simplifies the registration problem to a 3-DOF search over the longitudinal \(x\), lateral \(y\) and head rotation \(\theta\) of the vehicle pose. Here, we propose to maximize the normalized mutual information (NMI) \cite{Studholme1999} between reflectivity of candidate patches from the prior map \(M\) and the local grid-map \(L_i\) \cite{wolcott2014visual}.\\

The optimization problem that we solve is given below, 

\begin{equation}
    (\hat{x},\hat{y},\hat{\theta}) = \argmax_{x \in \mathcal{X},y \in \mathcal{Y},\theta \in \mathcal{H}}NMI (M, L_i) \label{search} \\[20pt]
\end{equation}
 where \(\mathcal{X}\), \(\mathcal{Y}\) and \(\mathcal{H}\) represent our search space in all 3 degrees of freedom and
\begin{equation}
	\text{NMI} (A, B ) = \frac{ H(A) + H(B) }{H(A,B)}.
\end{equation}
 $H(A)$ and $H(B)$ is the entropy of random variable $A$ and $B$, respectively and $H(A,B)$ is the joint entropy. Using NMI over standard mutual information has the advantages of being robust under occlusions \cite{Dame11}, less sensitive to the amount of overlap \cite{Studholme1999} while also maintaining the desirable properties of mutual information registration including outlier robustness \cite{Viola95}.

\subsection{Localization Filter}
\label{ssec:filter}

Our approach uses an Extended Kalman Filter (EKF) as the localization filtering framework. The state vector is represented by \begin{math} \mu = [x, y, \theta] \end{math}. The EKF localization framework consists of two main steps: \(Predict\) and \(Update\).
The prediction step involves predicting the motion of the vehicle between two iterations of the filter. We obtain velocities from the Applanix POS LV\cite{applanix} and apply a constant velocity motion model to estimate the pose of the vehicle (also called predicted state).
The update step is where the state of the filter is updated using measurements. The measurements are the output candidate of the NMI registration search based on \eqref{search}. These measurements correct the vehicle pose in \(x\), \(y\) and \(\theta\). Mathematically, this fusion is performed via the iterative localization updates

 \begin{alignat}{2} \label{EKF}
	\text{Predict:} & \quad \bar{\mu}_k = F_{k-1} \mu_{k-1} \\
	& \quad \bar{\Sigma}_k = F_{k-1} \Sigma_{k-1} F_{k-1}^T + Q_{k-1} \nonumber \\
	\nonumber \text{Update:} & \quad K_k = \bar{\Sigma}_k H^T_k ( H_k \bar{\Sigma}_k H^T_k + R_k)^{-1} \\
	\nonumber & \quad \mu_k =  \bar{\mu}_k + K_k(z_k - h_k( \bar{\mu}_k )) \\
	\nonumber & \Sigma_k = (I - K_k H_k) \bar{\Sigma}_k (I - K_k H_k)^T + K_k R_k K_k^T
 \end{alignat}
 
where $F_k$ represents the motion model of the vehicle and $Q_k$ is the corresponding uncertainty, $z_k$ is the output of the NMI registration in \eqref{search} and $R_k$ is the corresponding uncertainty estimated as a fit to the covariance of the NMI cost as was done in \cite{Olson09}. We use GPS to initialize the filter in the linearized global frame which results in high uncertainty for the first few measurements. The NMI registration is performed with a dynamic bound exhaustive search updated to a 3$\sigma$ window around the posterior distribution of the EKF. 

\section{Experimental Setup}
\label{Sec:Experiments}

In this section, we present the data collection vehicle, test dataset and various post processing techniques that were required to generate the performance results. 

\subsection{Platform}
\label{ssec:platform}
\begin{figure}[h]
	\centering
	\captionsetup{justification=centering}
	\subfloat{\includegraphics[width=0.9\linewidth]{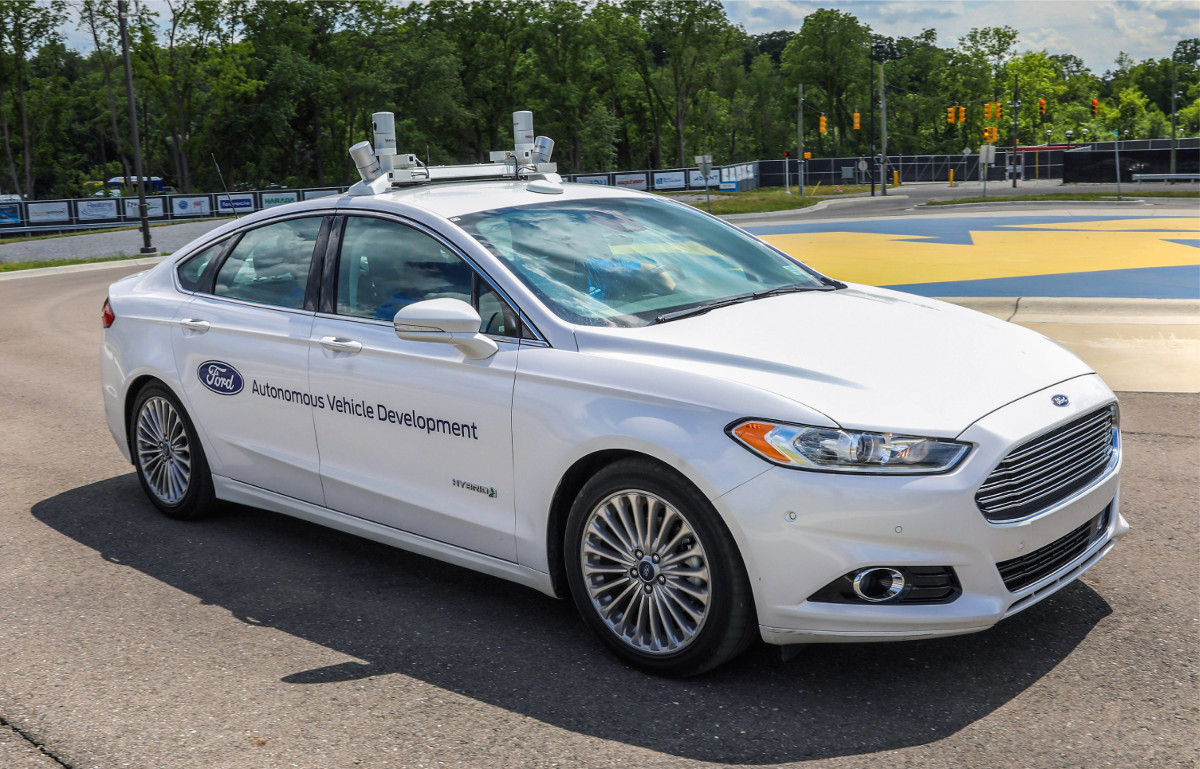}}
	\caption{Ford Fusion Autonomous Vehicle}
	\label{fig:fusion}
\end{figure}

The data was collected with a Ford Fusion test vehicle. This vehicle is fitted with four Velodyne HDL-32E 3D-LIDAR scanners and an Applanix POS-LV positioning system. All four LIDARs are mounted on the roof of the vehicle as shown in \ref{fig:fusion}; two of which rotate in an axis perpendicular to the ground and the other two are canted. The extrinsic calibration of the LIDARs is performed using GICP \cite{segal2009generalized}. 

\subsection{Dataset}
\label{ssec:dataset}
Mcity is a state-of-the-art testing facility designed for robust testing of autonomous vehicles. It simulates the broad range of complexities a vehicle could encounter in urban and highway environments. Among other urban attributes, the site contains railroad crossing, vegetation canopy, roundabouts, simulated buildings, and construction barriers\cite{mcity}. One 2.5 km long trajectory was logged at Mcity to determine the feasibility of our approach, as shown in Fig.\ref{fig:errorTrajectories}.

\subsection{Ground Truth Pose}
\label{ssec:groundTruth}
In this work we use full post-processed SLAM corrected pose of our datasets as ground-truth which is commonly used in the literature as an alternative to obtain a quantitative measure of error\cite{Wolcott17,castorena2017ground, wolcott2014visual,Levinson10}.

\subsection{LIDAR Ground Reflectivity Localization}
\label{ssec:IntensityLocalization}

 We compare our results with an implementation of the state-of-the-art localization technique by Levinson et al \cite{Levinson.Thrun2007}. We first create a ground reflectivity prior map followed by online matching with LIDAR local maps using NMI. The global LIDAR reflectivity prior map is obtained by estimating the full posterior of a pose-graph SLAM \cite{Durrant-Whyte.Bailey2006, Bailey.Durrant-Whyte2006}. The pose-graph is created with odometry, 3D LIDAR scan matching and GPS constraints. Here, the LIDAR constraints are imposed via generalized iterative closest point (GICP)\cite{segal2009generalized} and the back-end SLAM optimization is performed by incremental smoothing and mapping (iSAM)\cite{Kaess.Ranganathan.Dellaert2008}. From the optimized pose-graph, we create a dense ground plane and a full 3D point cloud which includes buildings, vegetation, road signs etc. by accumulating points in corresponding cells of a 3D grid.

\section{Analysis}
\label{sec:analysis}

To evaluate our localization performance, we compute errors in the lateral and longitudinal direction with respect to the vehicle. This error measures the performance of our online localization output to the offline (full) bundle adjustment corrected trajectory one would obtain with odometry and 3-D LIDAR matching constraints (section \ref{ssec:groundTruth}). In the same way, we also compute the error of ground reflectivity localization mentioned in section \ref{ssec:IntensityLocalization}.

\begin{figure*}[ht]
    \centering
	\subfloat[AL localization trajectory]{\includegraphics[width=0.375\linewidth, trim=0 90 0 0, clip]{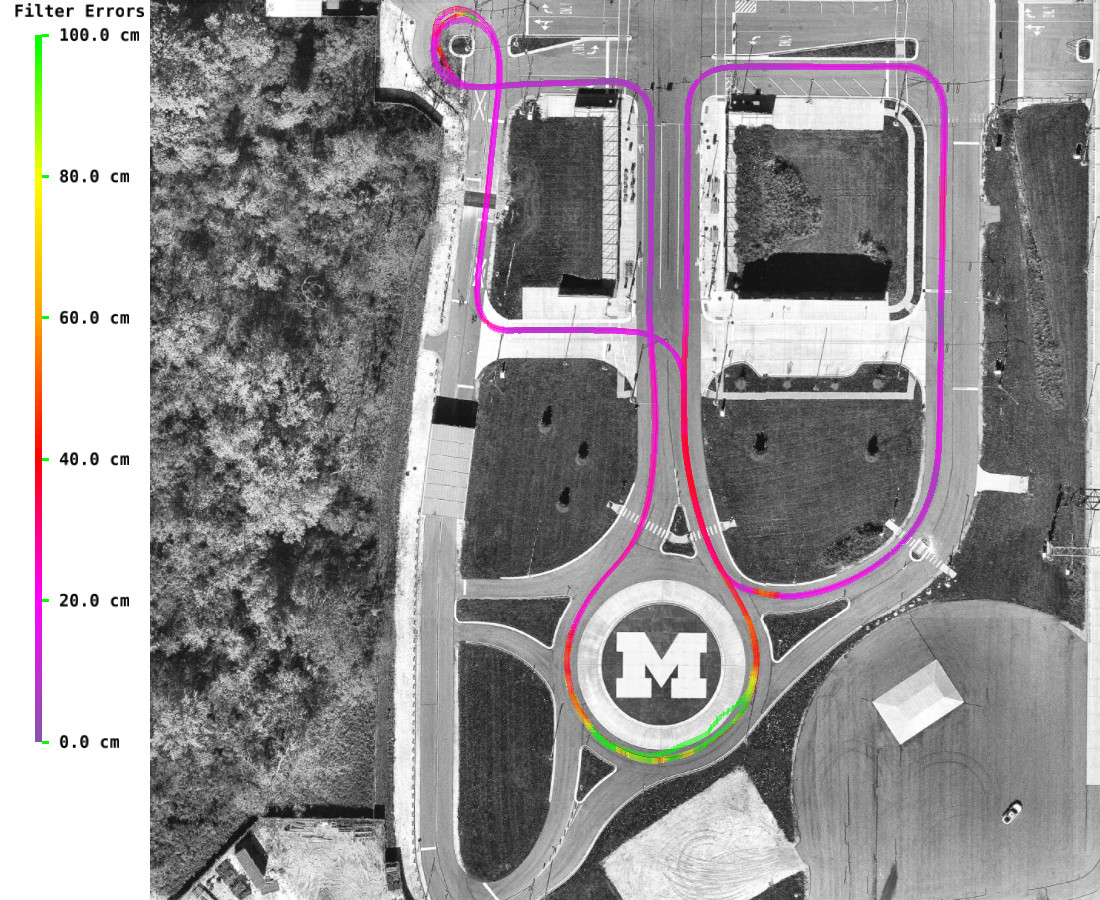}}
	\subfloat[LGR localization trajectory]{\includegraphics[width=0.316\linewidth,  trim=0 90 0 0, clip]{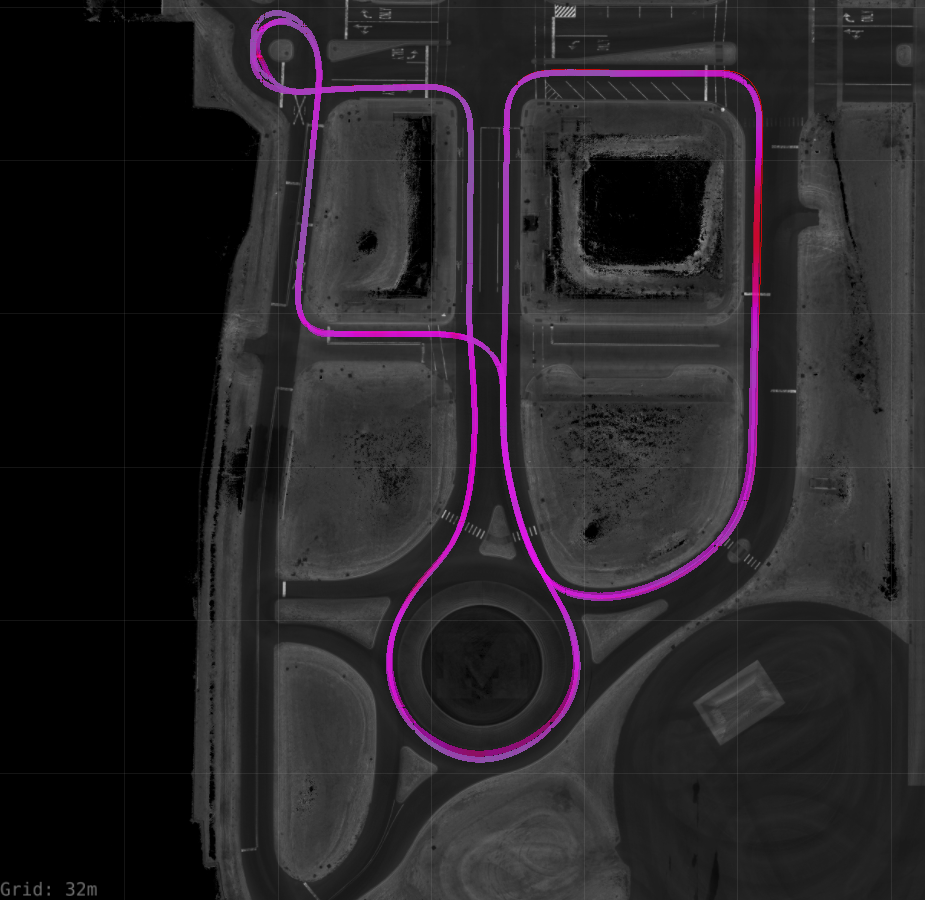}}
	\subfloat[Odometry Pose trajectory]{\includegraphics[width=0.28\linewidth,  trim=0 90 0 0, clip]{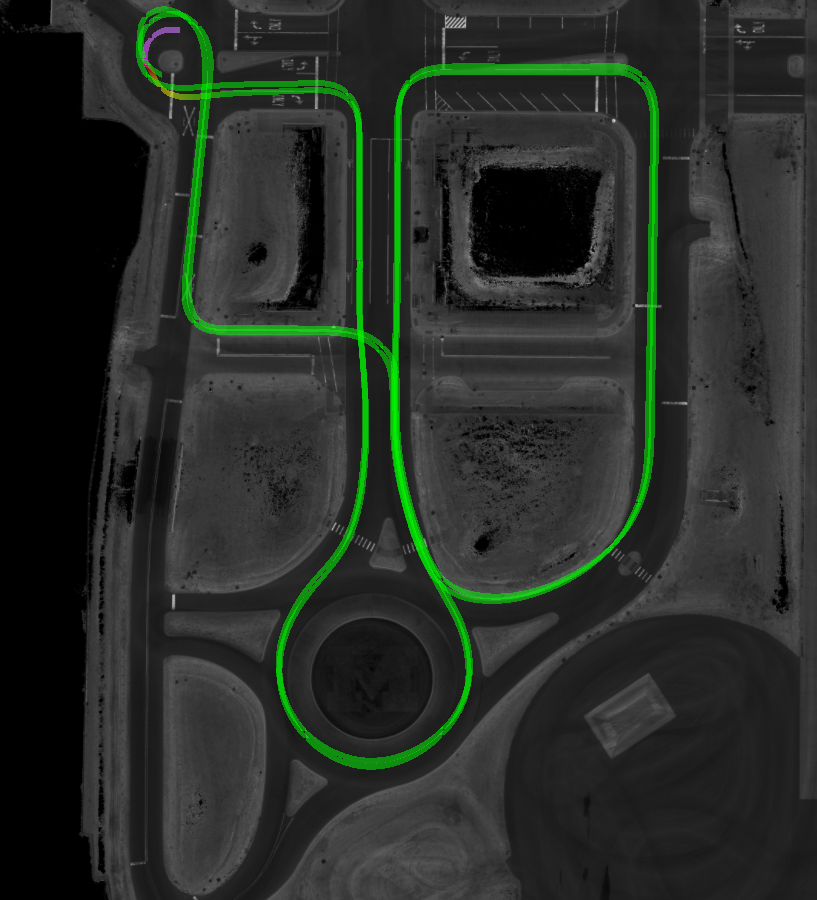}}
	\caption{Visual comparison between the trajectories from AL and LGR localization techniques along with Applanix odometry.}
	\label{fig:errorTrajectories}
\end{figure*}

\begin{figure}[h]
	\centering
	\subfloat{\includegraphics[width=0.81\linewidth]{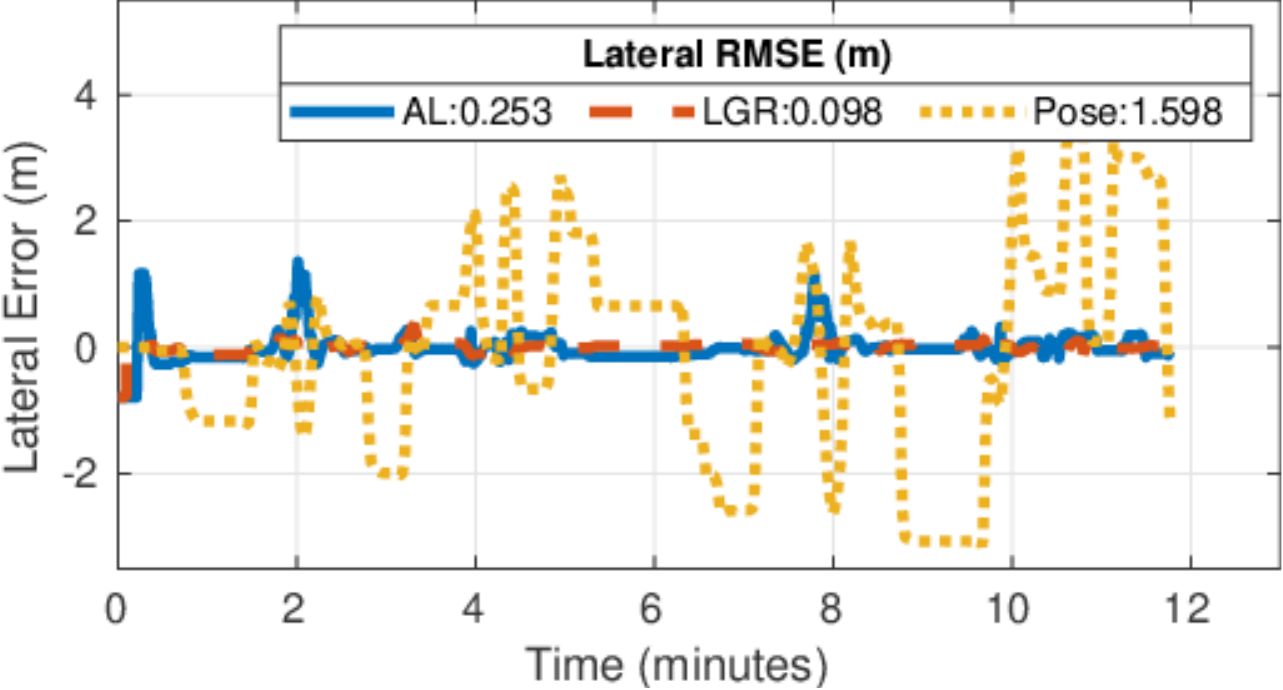}}\\
	\subfloat{\includegraphics[width=0.81\linewidth]{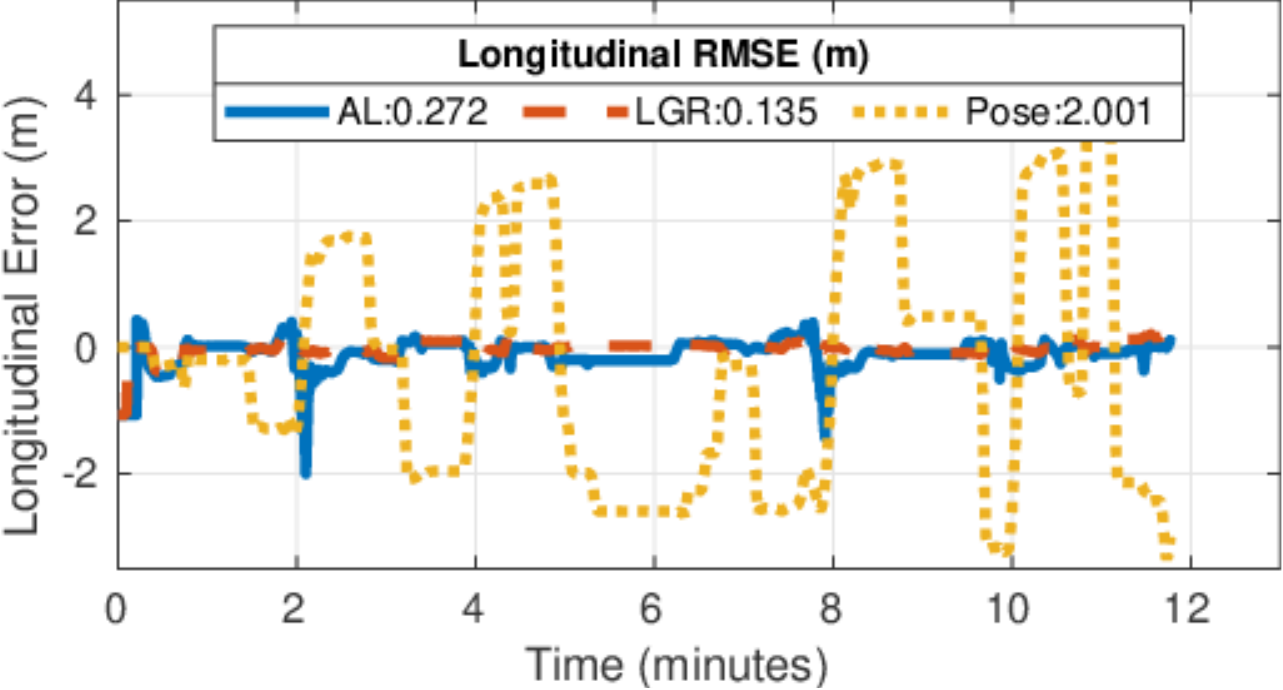}}\\
	\caption{Localization Error Comparisons}
	\label{fig:DenseLocError}
\end{figure}

Fig.~\ref{fig:DenseLocError} shows the performance of both AL and LGR techniques on the dataset. We also plot the odometry pose error (\(Pose\)) to show how it grows if a vehicle just relies on a GPS+IMU fusion system. Our aerial imagery based approach has a lateral and longitudinal error RMSE value of 0.253 m and 0.272 m as compared to 0.098 m and 0.135 m for the ground reflectivity based approach. In addition, our error is an order of magnitude lesser than a system that uses GPS+IMU based localization (1.598 m and 2.001 cm).

\begin{figure}
    \centering\
	\subfloat{\includegraphics[width=0.9\linewidth, trim=0 0 0 20]{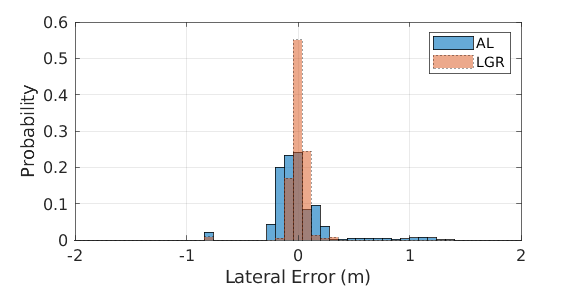}}\\
	\subfloat{\includegraphics[width=0.9\linewidth, trim=0 0 0 20]{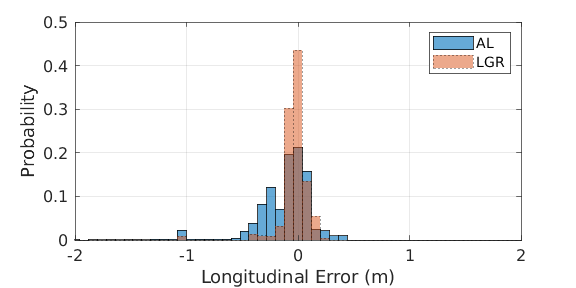}}
	\caption{Localization Error Histograms}
	\label{fig:errorHistogram}
\end{figure}

Fig.~\ref{fig:errorTrajectories} shows trajectory plots of AL, LGR and odometry pose. In fig.~\ref{fig:errorTrajectories}a one can note that the L2 error for AL goes over 1m at the roundabouts. In general NMI can deal with partial occlusions caused by shadows and objects, but in this case the grayscale color of the features at the roundabout are very different from what the LIDAR sees. As a result, the NMI produces incorrect matches until it encounters more recognizable features like the crosswalks, lane lines which are white both in the prior map as well as the local map. On the contrary, in fig.\ref{fig:errorTrajectories}b the ground reflectivity localization has no problems at the same location since the prior map and local map are both derived from the same source - LIDAR.

Our nominal operating map resolution is 8cm as mentioned in section \ref{ssec:globalMap}. Such high resolution imagery maybe difficult/expensive to obtain. Hence, we demonstrate the results of the technique with lower resolution maps (16cm \& 32cm). In order to produce the map at these resolutions, we down sample each tile image to 400 x 400 and 200 x 200 pixels respectively which ensures consistency with 64m x 64m tile size. To maintain consistency between local and prior map, we create local maps of same resolution for each experiment. Table.~\ref{tab:mapResolution} compares the localization performance results of our approach with varying resolutions and also against the LGR technique at those resolutions. Note that the error does not change much with degrading levels of map resolution for both AL and LGR.

\begin{table}[h]
	\caption{\label{tab:mapResolution}Localization performance}
	\centering
	\begin{tabular}{ccccccc}
		\hline
		 & \multicolumn{2}{c}{AL} & \multicolumn{2}{c}{LGR} & \multicolumn{2}{c}{Pose}\\
		Resolution & Lat & Long & Lat & Long & Lat & Long\\
		 (m) & (m) & (m) & (m) & (m) & (m) & (m)\\
		\hline
		 8 & 0.253 & 0.272 & 0.098 & 0.135 & 1.598 & 2.001\\
		 16 & 0.310 & 0.238 & 0.101 & 0.133 & 1.598 & 2.001\\
		 32 & 0.323 & 0.241 & 0.111 & 0.129 & 1.598 & 2.001\\
		\hline
	\end{tabular}
\end{table}

Reid et al.\cite{Reid_2019} established localization alert limits of 0.29 m for autonomous operation on local US roads. The errors in our method with 8 cm aerial maps are well within the limits 93.27\%(lateral) and 81.40\%(longitudinal) times for US local roads as can be seen in fig.~\ref{fig:errorHistogram}. Similarly for 32 cm maps, mimicking satellite imagery, we meet the requirements 89.44\%(lateral) and 84.91\%(longitudinal) times for US local roads. As with all techniques, we do experience some outliers in as shown in  Fig.~\ref{fig:errorTrajectories}. The reflectivity from an aerial/satellite image in some areas can be very different from that observed in local LIDARs scans, thus momentarily increasing the error and paving the path for future work.

\section{CONCLUSIONS}

In this investigation we demonstrate a localization technique using aerial imagery and LIDAR based local maps for autonomous vehicles in urban environments. Our approach works in real time and eliminates the requirement of producing computationally expensive LIDAR maps while achieving feasible performance with respect to the state-of-the-art LIDAR ground reflectivity based maps. This represents a significant advantage if one considers the cost of map data collection and the resources needed to maintain and update the existing maps. The aerial imagery acquired from a satellite or a low flying aircrafts provides an easier and more cost effective solution for building and maintaining maps at large scale.

\section*{ACKNOWLEDGMENT}

The authors thank Ford AV LLC and Ford Motor Company for their support towards this work. We also thank Kevin Bullock and his team at Maxar Technologies for their contribution of aerial imagery for these experiments.


\bibliographystyle{IEEEtranS}
\bibliography{refs}

\begin{thebibliography}{10}
\providecommand{\url}[1]{#1}
\csname url@samestyle\endcsname
\providecommand{\newblock}{\relax}
\providecommand{\bibinfo}[2]{#2}
\providecommand{\BIBentrySTDinterwordspacing}{\spaceskip=0pt\relax}
\providecommand{\BIBentryALTinterwordstretchfactor}{4}
\providecommand{\BIBentryALTinterwordspacing}{\spaceskip=\fontdimen2\font plus
\BIBentryALTinterwordstretchfactor\fontdimen3\font minus
  \fontdimen4\font\relax}
\providecommand{\BIBforeignlanguage}[2]{{%
\expandafter\ifx\csname l@#1\endcsname\relax
\typeout{** WARNING: IEEEtranS.bst: No hyphenation pattern has been}%
\typeout{** loaded for the language `#1'. Using the pattern for}%
\typeout{** the default language instead.}%
\else
\language=\csname l@#1\endcsname
\fi
#2}}
\providecommand{\BIBdecl}{\relax}
\BIBdecl

\bibitem{mcity}
\BIBentryALTinterwordspacing
``Mcity test facility.'' [Online]. Available:
  \url{https://mcity.umich.edu/our-work/mcity-test-facility/}
\BIBentrySTDinterwordspacing

\bibitem{swiftnav}
\BIBentryALTinterwordspacing
``Swift nav.'' [Online]. Available:
  \url{https://www.swiftnav.com/sites/default/files/whitepapers/localization_white_paper_052617.pdf}
\BIBentrySTDinterwordspacing

\bibitem{applanix}
\BIBentryALTinterwordspacing
Applanix. (2016) {Applanix POS LV}. [Online]. Available:
  \url{https://www.applanix.com/products/poslv.htm}
\BIBentrySTDinterwordspacing

\bibitem{slamBosse}
M.~Bosse, P.~Newman, J.~Leonard, and S.~Teller, ``Simultaneous localization and
  map building in large-scale cyclic environments using the atlas framework,''
  \emph{The International Journal of Robotics Research}, vol.~23, no.~12, pp.
  1113--1139, 2004.

\bibitem{castorena2017ground}
J.~Castorena and S.~Agarwal, ``Ground edge based lidar localization without a
  reflectivity calibration for autonomous driving,'' \emph{IEEE Robotics and
  Automation Letters}, 2017.

\bibitem{Veronese}
L.~d.~P.~{Veronese}, E.~{de Aguiar}, R.~C. {Nascimento}, J.~{Guivant}, F.~A.~A.
  {Cheein}, A.~F. {De Souza}, and T.~{Oliveira-Santos}, ``Re-emission and
  satellite aerial maps applied to vehicle localization on urban
  environments,'' in \emph{2015 IEEE/RSJ International Conference on
  Intelligent Robots and Systems (IROS)}, Sep. 2015, pp. 4285--4290.

\bibitem{Dame11}
A.~Dame and E.~Marchand, ``Mutual information-based visual servoing,''
  \emph{IEEE Transactions on Robotics}, vol.~5, no.~27, pp. 958--969, 2011.

\bibitem{dogruer2010outdoor}
C.~U. Dogruer, A.~B. Koku, and M.~Dolen, ``Outdoor mapping and localization
  using satellite images,'' \emph{Robotica}, vol.~28, no.~7, pp. 1001--1012,
  2010.

\bibitem{Durrant-Whyte.Bailey2006}
H.~Durrant-Whyte and T.~Bailey, ``Simultaneous localization and mapping: Part
  i,'' \emph{IEEE Robotics and Automation Magazine}, vol.~13, no.~2, pp.
  99--110, 2006.

\bibitem{Bailey.Durrant-Whyte2006}
------, ``Simultaneous localization and mapping: Part ii,'' \emph{IEEE Robotics
  and Automation Magazine}, vol.~13, no.~3, pp. 108--117, 2006.

\bibitem{FoxAMCL}
D.~Fox, W.~Burgard, F.~Dellaert, and S.~Thrun, ``Monte carlo localization:
  Efficient position estimation for mobile robots,'' 01 1999, pp. 343--349.

\bibitem{Kaess.Ranganathan.Dellaert2008}
M.~Kaess, A.~Ranganathan, and F.~Dellaert, ``isam: Incremental smoothing and
  mapping,'' \emph{IEEE Transactions on Robotics}, vol.~24, no.~6, pp.
  1365--1378, 2008.

\bibitem{kume2015bundle}
H.~Kume, T.~Sato, and N.~Yokoya, ``Bundle adjustment using aerial images with
  two-stage geometric verification,'' \emph{Computer Vision and Image
  Understanding}, vol. 138, pp. 74--84, 2015.

\bibitem{Levinson.Thrun2007}
J.~Levinson, M.~M., and S.~Thrun, ``Map-based precision vehicle localization in
  urban environments,'' in \emph{Robotics Science and Systems}, 2007.

\bibitem{Levinson10}
J.~Levinson and S.~Thrun, ``Robust vehicle localization in urban environments
  using probabilistic maps,'' \emph{IEEE International Conference on Robotics
  and Automation}, pp. 4372--4378, May 2010.

\bibitem{Levinson.Thrun2010}
------, ``Unsupervised calibration for multi-beam lasers,'' in
  \emph{International Symposium on Experimental Robotics}, 2010.

\bibitem{DARPA}
K.~I. M.~Buehler and S.~Singh, \emph{The 2005 DARPA Grand Challenge: The Great
  Robot Race}.\hskip 1em plus 0.5em minus 0.4em\relax Springer Publishing
  Company, 2007.

\bibitem{MaesImageRegistration}
F.~{Maes}, A.~{Collignon}, D.~{Vandermeulen}, G.~{Marchal}, and P.~{Suetens},
  ``Multimodality image registration by maximization of mutual information,''
  \emph{IEEE Transactions on Medical Imaging}, vol.~16, no.~2, pp. 187--198,
  April 1997.

\bibitem{ndjeng2011low}
A.~N. Ndjeng, D.~Gruyer, S.~Glaser, and A.~Lambert, ``Low cost
  imu--odometer--gps ego localization for unusual maneuvers,''
  \emph{Information Fusion}, vol.~12, no.~4, pp. 264--274, 2011.

\bibitem{nearmap}
\BIBentryALTinterwordspacing
Nearmap. (2019). [Online]. Available: \url{https://www.nearmap.com}
\BIBentrySTDinterwordspacing

\bibitem{nixon2013systems}
S.~W. Nixon, ``Systems and methods of capturing large area images in detail
  including cascaded cameras and/or calibration features,'' Jul.~30 2013, uS
  Patent 8,497,905.

\bibitem{NodaInVehicle}
M.~Noda, T.~Takahashi, D.~Deguchi, I.~Ide, H.~Murase, Y.~Kojima, and T.~Naito,
  ``Vehicle ego-localization by matching in-vehicle camera images to an aerial
  image,'' in \emph{Computer Vision -- ACCV 2010 Workshops}, R.~Koch and
  F.~Huang, Eds.\hskip 1em plus 0.5em minus 0.4em\relax Springer Berlin
  Heidelberg, 2011, pp. 163--173.

\bibitem{obst2012multipath}
M.~Obst, S.~Bauer, P.~Reisdorf, and G.~Wanielik, ``Multipath detection with 3d
  digital maps for robust multi-constellation gnss/ins vehicle localization in
  urban areas,'' in \emph{2012 IEEE Intelligent Vehicles Symposium}.

\bibitem{Olson09}
E.~Olson, ``Real-time correlative scan matching,'' in \emph{IEEE International
  Conference in Robotics and Automation}, Kobe, Japan, June 2009.

\bibitem{ortho}
\BIBentryALTinterwordspacing
OSSIM. (2014). [Online]. Available:
  \url{https://trac.osgeo.org/ossim/wiki/orthorectification}
\BIBentrySTDinterwordspacing

\bibitem{ranganathan}
A.~Ranganathan, D.~Ilstrup, and T.~Wu, ``Light-weight localization for vehicles
  using road markings,'' 11 2013, pp. 921--927.

\bibitem{Reid_2019}
\BIBentryALTinterwordspacing
T.~G. Reid, S.~E. Houts, R.~Cammarata, G.~Mills, S.~Agarwal, A.~Vora, and
  G.~Pandey, ``Localization requirements for autonomous vehicles,'' \emph{SAE
  International Journal of Connected and Automated Vehicles}, vol.~2, no.~3,
  Sep 2019. [Online]. Available: \url{http://dx.doi.org/10.4271/12-02-03-0012}
\BIBentrySTDinterwordspacing

\bibitem{Rohde}
J.~{Rohde}, I.~{Jatzkowski}, H.~{Mielenz}, and J.~M. {Zöllner}, ``Vehicle pose
  estimation in cluttered urban environments using multilayer adaptive monte
  carlo localization,'' in \emph{2016 19th International Conference on
  Information Fusion (FUSION)}, July 2016, pp. 1774--1779.

\bibitem{segal2009generalized}
A.~Segal, D.~Haehnel, and S.~Thrun, ``Generalized-icp.'' in \emph{Robotics:
  science and systems}, vol.~2, no.~4.\hskip 1em plus 0.5em minus 0.4em\relax
  Seattle, WA, 2009, p. 435.

\bibitem{SenletRobotAerial}
T.~{Senlet} and A.~{Elgammal}, ``Satellite image based precise robot
  localization on sidewalks,'' in \emph{2012 IEEE International Conference on
  Robotics and Automation}, May 2012, pp. 2647--2653.

\bibitem{Studholme1999}
C.~Studholme, D.~Hill, and D.~Hawkes, ``An overlap invariant entropy measure of
  3d medical image alignment,'' \emph{Pattern Recognition}, vol.~32, no.~1, pp.
  71--86, January 1999.

\bibitem{maxar}
\BIBentryALTinterwordspacing
M.~Technologies. (2020). [Online]. Available: \url{https://www.maxar.com/}
\BIBentrySTDinterwordspacing

\bibitem{slamThrun}
S.~Thrun and M.~Montemerlo, ``The graph slam algorithm with applications to
  large-scale mapping of urban structures,'' \emph{The International Journal of
  Robotics Research}, vol.~25, no. 5-6, pp. 403--429, 2006.

\bibitem{Viola95}
P.~Viola, ``Alignment by maximization of mutual information,'' Ph.D.
  dissertation, Massachusetts Institute of Technology, 1995.

\bibitem{Wolcott17}
R.~Wolcott, D.~Wollherr, and M.~Buss, ``Robust lidar localization using
  multiresolution gaussian mixture maps for autonomous driving,'' \emph{The
  Int. Journal of Robotics Research}, vol.~36, no.~3, pp. 292--319, 2017.

\bibitem{wolcott2014visual}
R.~W. Wolcott and R.~M. Eustice, ``Visual localization within lidar maps for
  automated urban driving,'' in \emph{Intelligent Robots and Systems (IROS
  2014), 2014 IEEE/RSJ International Conference on}.\hskip 1em plus 0.5em minus
  0.4em\relax IEEE, 2014, pp. 176--183.

\end{thebibliography}


\end{document}